%% file: main.tex
\documentclass[10pt,twocolumn,letterpaper]{article}

\usepackage{iccv}
\usepackage{times}
\usepackage{epsfig}
\usepackage{graphicx}
\usepackage{amsmath}
\usepackage{subcaption}

\usepackage{amssymb}
\usepackage{xcolor}
\usepackage{ifthen}

\usepackage[pagebackref=true,breaklinks=true,colorlinks,bookmarks=false]{hyperref}

\iccvfinalcopy 

\newcommand{\fig}[1]{Fig.~\ref{#1}}

\begin{document}

\title{Eyemotion: Classifying facial expressions in VR using eye-tracking cameras}

\author{Steven Hickson\\
Georgia Institute of Technology\thanks{Work was done as part of a Google internship}\\
{\tt\small me@stevenhickson.com}
\and
Nick Dufour\\
Google\\
{\tt\small ndufour@google.com}
\and
Avneesh Sud\\
Google\\
{\tt\small avneesh@google.com}
\and
Vivek Kwatra\\
Google\\
{\tt\small kwatra@google.com}
\and
Irfan Essa\\
Georgia Institute of Technology\\
{\tt\small irfan@gatech.edu}
}

\maketitle

\begin{abstract}
One of the main challenges of social interaction in virtual reality settings is that head-mounted displays occlude a large portion of the face, blocking facial expressions and thereby restricting social engagement cues among users. Hence,  auxiliary means of sensing and conveying these expressions are needed. We present an algorithm to automatically infer expressions by analyzing only a partially occluded face while the user is engaged in a virtual reality experience. Specifically, we show that images of the user's eyes captured from an IR gaze-tracking camera within a VR headset are sufficient to infer a select subset of facial expressions without the use of any fixed external camera. Using these inferences, we can generate dynamic avatars in real-time which function as an expressive surrogate for the user. We propose a novel data collection pipeline as well as a novel approach for increasing CNN accuracy via personalization. Our results show a  mean accuracy of 74\% ($F1$ of 0.73) among 5 `emotive' expressions and a mean accuracy of 70\% ($F1$ of 0.68) among 10 distinct facial action units.
\end{abstract}

\input{intro}  
\input{related}
\input{data}

\input{approach}

\input{results}

\input{discussion}
\input{closing}

{\small
\bibliographystyle{ieee}
\bibliography{egbib}
}

\end{document}

%% file: intro.tex
\section{\label{sec:intro}Introduction}

\begin{figure}[t!]
	\centering
    \includegraphics[width=\columnwidth]{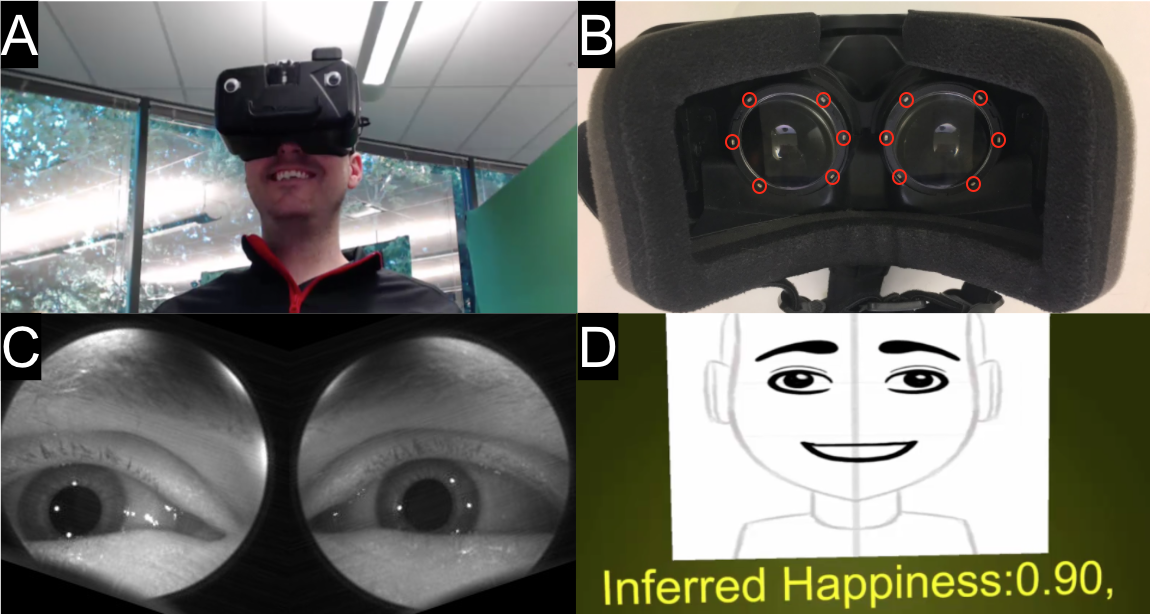}
    \caption{Eyemotion visual schematic. A: A user wearing the VR HMD used for expression tracking (Note that no external camera is used in our method; this is just for visualization). B: Interior of the HMD, with IR LEDs visible around the radius of the eyepieces, highlighted with red circles. C: Captured eye data. D: Model inference with dynamic avatar representation.}
    \label{fig:vr-headset}
    \vspace{-.1in}
\end{figure}

Facial expressions are essential for interpersonal communication and social interaction. They provide a means for conveying thought and emotion through visual cues that may not be easy to articulate verbally. However, virtual reality (VR) equipment using head-mounted displays (HMD) makes natural expressions difficult to recognize as half the face is occluded. Thus for VR systems to provide rich social interaction, faithfully representing these expressions in some manner is absolutely critical. We propose to recognize and convey facial expressions from inside a VR HMD.

Visual classification of expressions has been a well studied topic in computer vision. However, most of this work is aimed at classification with a fixed front-facing camera that relies on seeing the full face of a user~\cite{fasel2003automatic, liu2014facial, saatci2006cascaded}. We focus on a more challenging scenario with no fixed external camera, where the user is wearing a head-mounted display (HMD) in a VR setting as shown in (\fig{fig:vr-headset}A). 

One could attach a user facing external camera that captures the lower face but that may not always be feasible such as in a mobile setting. An external camera is also not able to capture upper-facial expressions as they are occluded. We propose a new approach aimed at classification of facial action units (AUs)~\cite{ekman1977facial} and `emotive' expressions using only internally mounted infrared cameras within the HMD.

We are motivated by the recent availability of commercial HMDs with eye-tracking cameras~\cite{SMI}. To avoid interference with the VR display, infrared cameras are mounted behind the lens, and point at the eyes and surrounding areas (\fig{fig:vr-headset}B). The images are typically used for eye-gaze estimation and for applications such as foveated rendering~\cite{Patney_2016}, but in our work we use the same input images for expression classification. A key aspect of our work is a \emph{labeling}-based approach -- as opposed to visual tracking of facial features -- to identify expressions and action units only from these  limited field-of-view \emph{eye images}. 

Our model classifies user expressions using only limited periocular eye image data, as shown in \fig{fig:vr-headset}C, which is further limited by the large amount of intra-class variation among users. To account for limited data, a new type of data, and large variations, we turn to deep learning techniques. Recently convolutional neural networks (CNNs)~\cite{krizhevsky2012imagenet, girshick2014rich, simonyan2014very} have performed very well on image classification tasks and are pervasive in machine learning and computer vision. Additionally, deep learning methods have the benefit of learning important invariant features and embeddings without requiring any hand-crafted feature representations. Deep learning has also been shown to give state of the art results on faces~\cite{li2015facial,schroff2015facenet}.
Our approach, based on deep learning, outperforms normal human accuracy and even advanced (trained users) human accuracy for categorizing select facial expressions from our dataset of only IR eye images. Human ratings form the primary baseline for our work. We use these ratings for comparison and evaluation, but \emph{not} as labels during training.

We also demonstrate an application of our classification framework that animates facial models and avatars in real-time, which could be used in social VR apps to convey and interpret users' facial expressions. Note that our \emph{classification} based approach has applications beyond synthesis since it also provides semantics of the expression (potentially correlated with the emotive response to the VR environment), which may then be used as feedback to the system.

\medskip
\noindent\textbf{Our primary contributions are:}
\textbf{(1)} Demonstrating that the information required to classify a variety of facial expressions is reliably present in IR eye images captured by a commercial HMD sensor, and that this information can be decoded using a CNN-based method.
\textbf{(2)} A novel personalization technique to improve CNN accuracy on new users. Across experiments, personalization resulted in a 7\% accuracy improvement on average, and was statistically significant for a set of basic `emotive' expressions ($p = 0.018$) and AUs ($p = 0.001$) (Section~\ref{sec:personalize}).
\textbf{(3)} The collection of a unique dataset (Section~\ref{sec:data}) of eye images paired with expression labels, collected with two separate commercial HMDs each with 23 different users.
\textbf{(4)} We show our method can be used to generate expressive avatars in real-time, which can function as an expressive surrogate for users engaged in VR environments  (Section~\ref{sec:applications}).


%% file: related.tex
\section{\label{sec:related}Related work}

The problem of expression inference from limited facial data in a VR setting exists at the intersection of well studied problems in a variety of disciplines, discussed below. 
\vspace{-.1in}
\paragraph{Expression classification from visual data:} While much work has been done on automatically inferring human expressions from images, nearly all of it focuses on unoccluded, frontal faces (see~\cite{Pantic:2000:AAF,fasel2003automatic,vinayface,sariyanidi2015automatic} for recent surveys). Tian {\em et al.}~\cite{Tian:2001:RAU} employ facial action units for fine-grained facial expression recognition, using feature tracking and neural networks. Saatci {\em et al.}~\cite{saatci2006cascaded} use active appearance models to construct cascaded SVM classifiers for gender and 4 expressions/emotional states. 
The popularity of deep learning produced renewed interest in the field, with new challenges like~\cite{dhall2016emotiw} requiring emotion recognition on video data, and~\cite{Benitez-Quiroz_2016_CVPR} requiring classification of 11 AUs and recognition of basic and compound emotions. Recent work on expression inference includes Liu {\em et al.}~\cite{liu2014facial}, who propose a boosted deep belief network, and Barsoum {\em et al.}~\cite{Barsoum_2016}, who use the VGG network~\cite{simonyan2014very} to learn emotions from noisy crowd-sourced labels. Kahou {\em et al.}~\cite{kahou2013combining} and Bargal {\em et al.}~\cite{Bargal_2016} use deep convolutional networks combined with SVM classifiers for expressions from the EmotiW dataset challenge~\cite{dhall2016emotiw}. Benitez-Quiroz {\em et al.}~\cite{Benitez-Quiroz_2016_CVPR} use  Kernel Subclass Discriminant Analysis (KSDA) to classify facial action units, intensities and emotion categories on a large image dataset in the wild. All of these works require \emph{full}, \emph{unoccluded} face images, unavailable in our scenario.
\vspace{-.1in}
\paragraph{Expression classification with alternate sensors:} There has been some recent research on expression classification using wearable sensors. Scheirer {\em et al.}~\cite{scheirer1999expression} use face mounted piezoelectric sensors to discriminate between confused and interested expressions as well as discriminating between expressive states in general and neutral states.
More recently Masai {\em et al.}~\cite{masai2016facial} used optical sensors mounted on glasses to determine 8 expressions similar to~\cite{dhall2016emotiw} with a small set of users, and performed hardware modifications for facial expression mapping inside a HMD~\cite{suzuki2016facial}.  This work is the closest to our proposal of expression classification in virtual reality headsets. However, we propose a method for expression classification using gaze tracking cameras rather than embedded optical or piezoelectric sensors, which is more robust to personalized fit, and does not \emph{require} customized sensors beyond a commercial eye-tracking system. Much work using alternate sensors has been motivated by facial re-enactment (\eg~\cite{li2015facial}) and is covered below.
\vspace{-.1in}
\paragraph{Gaze tracking in VR:} Gaze tracking is the subject of intense and sustained research, and used for many interactive applications~\cite{morimoto2005eye}. Gaze tracking has recently seen applications in virtual and mixed reality~\cite{lee2007robust,thies2016facevr} and on mobile devices using convolutional neural networks~\cite{cvpr2016_Khosla}.
However, these methods do not classify expressions or AUs.
\vspace{-.1in}
\paragraph{Facial re-enactment in VR:} 
Unoccluded face synthesis, facial reenactment and avatar re-targeting in VR has been an active area of recent research. These approaches use a combination of visual tracking of the lower (unoccluded) face, along with custom sensors mounted inside the HMD.
Burgos {\em et al.}~\cite{burgos2015real} composite the occluded part of the face by aligning and blending the unoccluded parts with a pre-captured facial expression database. Li {\em et al.}~\cite{li2015facial}  measure strain signals with electronic sensors to estimate facial expressions of the occluded face, and combine with a user facing RGB-D camera to train a linear 3D parametric model of blendshapes. Olszewski {\em et al.}~\cite{olszewski2016high} propose an approach for 3D avatar control with eye-tracking cameras and a user facing RGB camera. A CNN model is trained to regress from these streams to blendshape coefficients. Thies {\em et al.}~\cite{thies2016facevr} also perform real-time gaze-aware facial reenactment in VR using a RGB-D camera to capture the unoccluded regions, and two internal infra-red (IR) cameras to track the eye gaze. A multi-linear blendshape model is fit to these streams by optimizing for photometric and geometric alignment. Zhao {\em et al.}~\cite{zhao2016maskoff} synthesize unoccluded face images using an HMD case fitted with wide-angle near IR cameras, along with a user facing RGB camera.  A 3D bi-linear blendshape model is fit to the input, the occluded part is synthesized from warped, and colorized IR eye images with evaluation on unoccluded face images.

In contrast, our work aims to classify a set of facial expressions from cameras present in eye-tracking enabled VR HMDs, with all optical and field-of-view constraints. As such our work cannot be directly compared against these; instead we show comparisons among various CNN-based models trained on eye images and against human ratings  to benchmark our results. Most attempts at personalization use per-subject samples and quick retraining~\cite{castro2015predicting,zhao2016peak}. However there has also been some work at personalizing expression classification without retraining with new samples~\cite{ChuDC13} based on unsupervised generalization with STM. We however, introduce a novel personalization approach which requires no retraining with a deep learning framework. 

%% file: data.tex
\section{Our Dataset}\label{sec:data}

We perform supervised training using a CNN to classify face expressions from eye images recorded inside the HMD. We collect a large amount of data which is cumbersome or impossible to label by hand. This section describes our data collection method, which obtains ground truth expression data from participants using infra-red eye images and without the need for manual image annotations.

\subsection{Data Collection}

There exist public datasets of near infra-red eye images~\cite{bowyer16ndiris} that target iris detection and biometric authentication, but contain no expression labels. In addition, these datasets do not directly translate to the novel sensors used in VR. Therefore, we designed a system to collect such data from participants in a controlled setting.

\begin{figure}[t!]
	\centering
	\includegraphics[width=1\columnwidth]{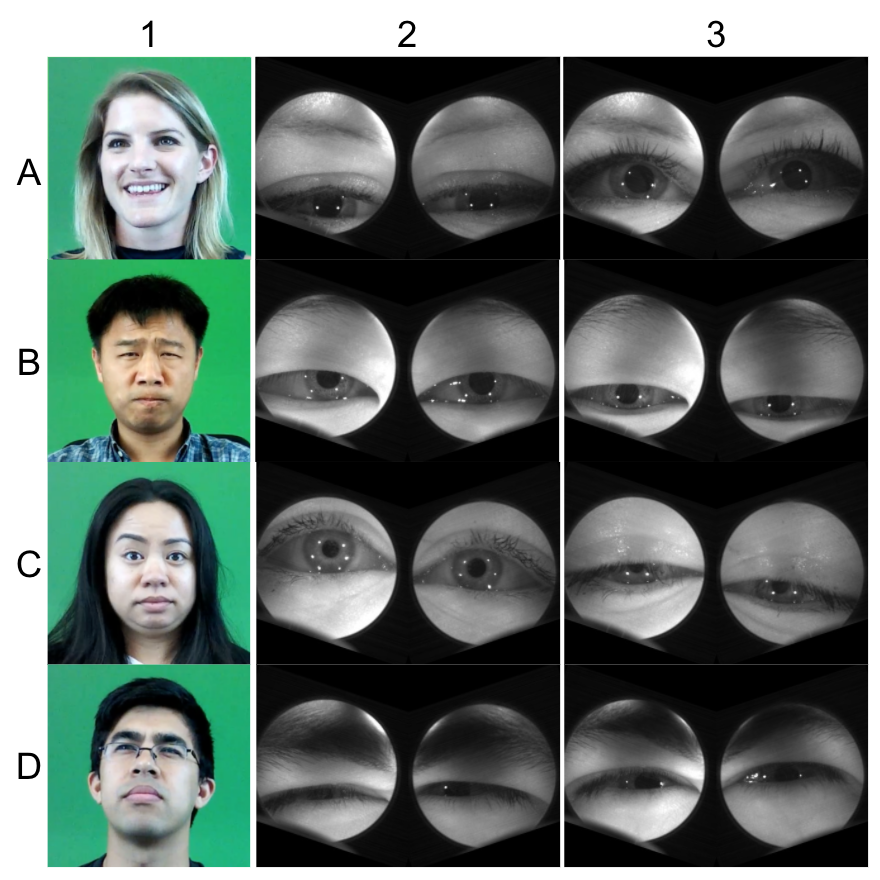}
	\caption{Inferring expressions from eye images alone is significantly different from doing so using the unoccluded face. Here we have four participants making four different expressions: without wearing the HMD (column 1, for reference only), within the HMD (column 2), and within the HMD during session 2 (column 3). The expressions are (A) happiness, (B) anger, (C) surprise, (D) and squint. The difficulty of this task, even for humans, is clear as the amount of expressive information conveyed is significantly reduced while variability is readily visible.}
\label{fig:sess_var}
\vspace{-.2in}
\end{figure}

We collected a subset of facial action units that influence the upper face, and could be reliably performed by multiple subjects. We also distinguish between left and right AUs, where applicable. These are \textsl{Neutral (AU0), Left Brow Raise (AU1+2L), Right Brow Raise (AU1+2R), Brow Lower (AU4), Upper Lid Raise (AU5), Squint (AU44), Both Eyes Closed (AU43), Left Wink (AU46L), Right Wink (AU46R),} and \textsl{Cheek Raise (AU6)}.
We also collect `emotive' expressions for basic emotions as defined by~\cite{ekman1977facial}, which are \textsl{Neutral, Anger, Surprise,} and \textsl{Happiness}. See \fig{fig:sess_var} for an illustration of the variability of the data. We experiment with training classifiers on facial action units, a subset of useful and non-overlapping facial action units, expressions, and a subset of expressions useful for VR social environments. The mapping from AUs to emotional expressions has been well characterized ~\cite{ekman1977facial}. 

\begin{figure*}[htp!]
	\centering
    \includegraphics[width=1.0\textwidth]{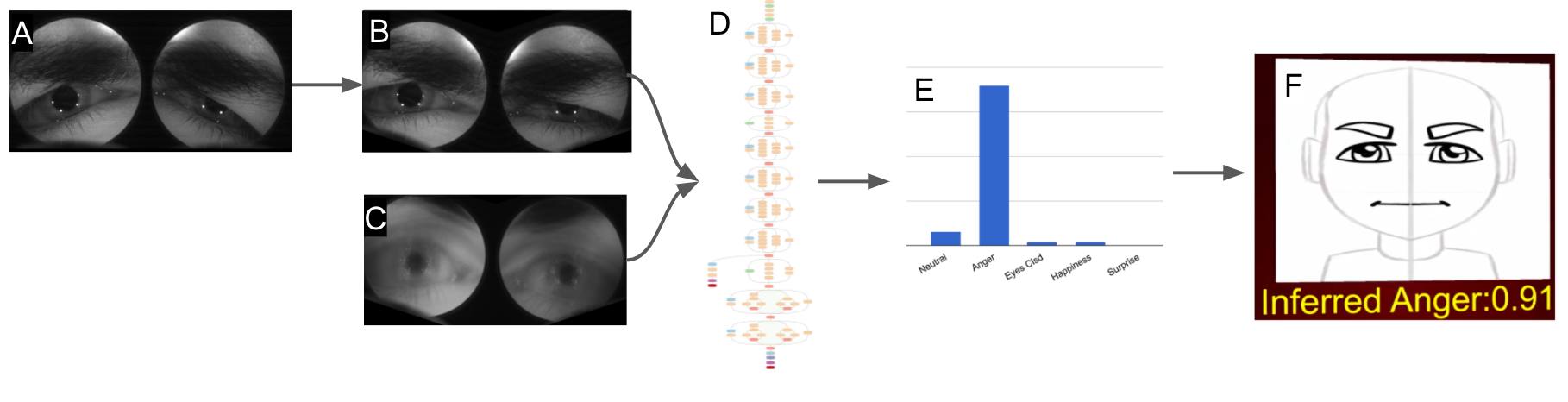}
    \caption{The Eyemotion pipeline, including personalization. A: Raw eye images from the HMD. B: Rectified eye images. C: The average neutral image for this user session, used for personalization. D: The difference between the rectified headset image and the mean neutral image is the input to a deep neural network. In the non personalization case, the mean neutral image is not subtracted from the rectified image. E: Output takes the form of a distribution over expressions. F: This distribution is used to generate an expressive avatar.}
    \label{fig:pipeline_v2}
    \vspace{-.1in}
\end{figure*}

\subsection{Collection Setup}
We collected data with two separate HMDs, with near IR (880nm) cameras mounted between the eye lens and display screen using a beam splitter. HMD1 and HMD2 capture 200x200-pixel and 320x240-pixel eye image pairs respectively, both at 10Hz.
The different optical properties of each HMD allow characterization of the generalization of our technique. 23 different participants were collected with each HMD with different genders, ethnicities, and hair color. 

We collected these data by asking users to form an expression, giving them an example from an exemplar video. While this may not result in spontaneous expressions~\cite{zhang2014bp4d}, it provides explicit labels for each expression. Removing the need for expert ratings allows larger-scale data collection than would otherwise be possible. To provide a realistic exemplar, we first recorded videos of trained actors performing each expression for the participant to use as a reference. During the collection process, for each expression, we provide to the participant the name of the expression, a looped clip of an actor performing the expression, and a live video of the participant in order for them to practice the expression. If the participant can't perform the expression, they are able to skip it. Otherwise they hold the expression and follow a randomly moving target on the screen with their gaze (to encourage gaze diversity) or head pose (to encourage variations in HMD pose).
This continues for all expressions and AUs (these are the images in column 1 of \fig{fig:sess_var}). We then have them put on the HMD and repeat the process twice more, taking the headset off and putting it back on to account for slippage and variation in fit. Each of these headset repetitions constitutes a `session.'

\subsection{Data Characteristics}
To ensure diversity, we collected each participant's personal information through a data collection form. Of the 46 participants: 16 were female; 16 were aged 35 or over from an age range of 18 to 64 with a median age of 30; 
11 participants had non-brown eyes and 4 had non-brown or black hair. 25 of our participants were nonwhite, with 9 Asian, 7 east Indian or south Asian, 4 two or more races, 3 Hispanic or Latino, 2 African American, and 2 preferring not to say. 

Approximately 50,000 eye image pairs were collected per HMD (about 2,000 per participant). Each expression was collected for the same amount of time; however, as some participants were unable to perform all of the expressions, there exists slight variability in the number of images per expression. 

\subsection{Data Cleanup}
Since participants blink during the data collection process, and we want to remove this source of variance, images with closed eyes were removed from non-eyes-closed expressions using a classifier we train, similar to the approach in Section~\ref{sec:cnn_architecture}. The classifier was trained to recognize eyeblinks using the neutral and the eyes closed images from our participants. 
Neutral and closed-eye images were validated and cleaned manually by ensuring that neutral images did not include any blinks and that eyes were actually closed during closed-eye periods. Approximately 400 images were removed by hand and using the classifier.

\subsection{Data Augmentation}\label{sec:data_aug}

The tightly controlled acquisition environment in the headset means lighting and eye-camera viewpoint are largely fixed, with nearly all of the variation coming from differences in participants and the position of the headset on the head during acquisition. Thus data augmentation was performed carefully and monitored for consistency. Further, the semantic meaning of many of the labels (\eg\ `Left Eye Wink') precludes random flips. Random rotations are limited so that they do not exceed human variation in eye orientation. We found a 2\% variation (rather than 10\% used frequently) is appropriate when performing random augmentation of rotation, scale, and brightness.

%% file: approach.tex
\section{Our Approach}\label{sec:approach}

\begin{figure}[h] 
\begin{subfigure}{0.22\textwidth}
	\centering 
    \includegraphics[width=0.8\textwidth]{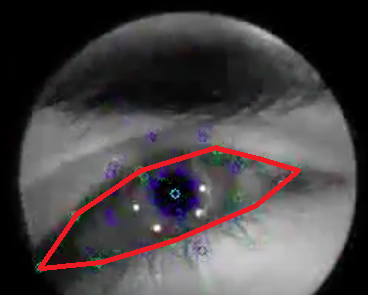}
	\caption{A correct example of the eye landmark tracker.}
	\label{fig:good_tracker}
\end{subfigure}
\begin{subfigure}{0.22\textwidth}
	\centering 
    \includegraphics[width=0.8\textwidth]{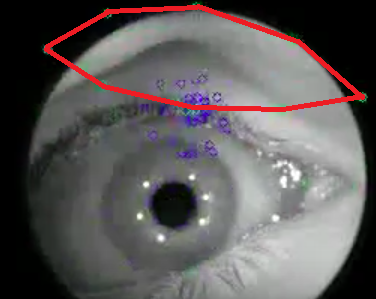}
	\caption{Most eye landmark tracker results contained errors.}
	\label{fig:bad_tracker}
\end{subfigure}
\caption{Eye tracker results with blue dots as the iris and pupil landmark locations and green dots as the palpebral fissure landmark locations.}
\label{fig:iris_tracker}
\vspace{-.1in}
\end{figure}
Initially, we experimented with features obtained from facial iris and eye landmark positions, similar to those produced in \cite{fasel2003automatic} and \cite{zhu2005robust}. However, testing with a proprietary eye landmarker produced frequent failures when the participants made expressions that distorted the shape of the eyes (see \fig{fig:iris_tracker}). Using those kinds of existing hand-crafted features on our new domain does not yield good features and ignores the periocular data which is found to contain useful information~\cite{park2011periocular}. Recent work on facial animation using VR and an external camera~\cite{li2015facial} has also shown that deep models work well to estimate facial expressions, which is similar to our problem. The entire pipeline, including personalization, is represented schematically in \fig{fig:pipeline_v2}.
	
\subsection{CNN Architecture}\label{sec:cnn_architecture}

Our proposed method leverages a CNN to learn an embedding describing expressions and emotions using infrared eye images. Specifically, we train a variant of the widespread Inception architecture~\cite{szegedy2015rethinking} using the TensorFlow library~\cite{abadi2016tensorflow} motivated by experimentation on various CNN architectures, described in Section~\ref{sec:results}. The model used was pre-trained for 150,000 iterations on the Imagenet data. Data are registered and augmented as described in Section \ref{sec:data}. The HMD eye cameras are calibrated, and both eye images are rectified, concatenated and scaled to 299x299 pixels. 
The network was trained using a learning rate of 0.045 which decays stepwise by 0.94 every epoch. To prevent overfitting an aggressive L2 weight decay was used (0.0004). A softmax cross-entropy (Eq~\ref{eq:softmax_ce}) function was chosen as the loss along with L2 regularization:

\begin{multline}
L(w) = - \frac{1}{N} \sum_{n=1}^{N} \sum_{c=1}^{C}  \left[ y_n^c \log{\hat{y}_n^c} - (1 - y_n^c) \log{(1 - \hat{y}_n^c)} \right] \\ +~\frac{\lambda}{N} \sum_{w}w^2 
\label{eq:softmax_ce}
\end{multline}
\noindent where $N$ is the number of samples, $C$ are the classes, $y_n^c$ and $\hat{y}_n^c$ are respectively the ground-truth label and softmax activation of the $c^\text{th}$ class for the $n^\text{th}$ sample, and $w$ are the network weights.

Optimization was performed with a `RMSProp' optimizer~\cite{tieleman2012lecture}, with momentum 0.9, decay factor 0.9, and $\epsilon$ of 1.0.

\subsection{Personalization}
\label{sec:personalize}
\begin{figure}[htp!]
	\centering
	\includegraphics[width=0.45\textwidth]{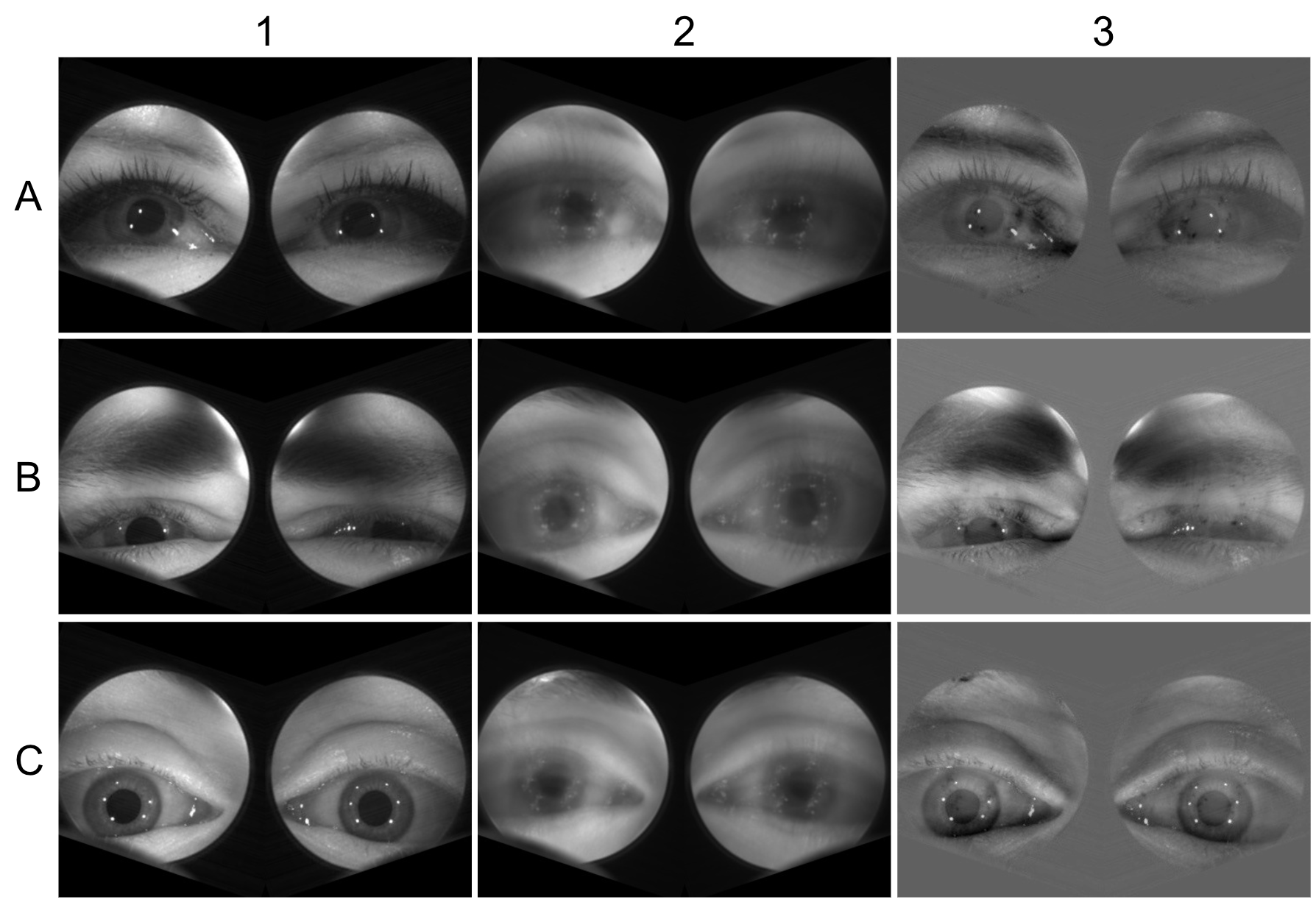}
	\caption{Personalization is performed by subtracting each user's mean neutral data from the current image to reduce the unimportant sources of variance and highlight important ones. Columns are 1: the original image, 2: the mean neutral image, and 3: the difference between the two (which are contrast normalized for demonstrative purposes). Row are different expressions, with A: happiness, B: anger, and C: brow raise.}
	\label{fig:personalization}
    \vspace{-.1in}
\end{figure}

One of the dominant sources of variance in our data is individual variation in appearance. We attempt to partially remove this variance since it does not vary with, and may not be predictive of, affective state. Since this variance removal occurs within-subject, it is effectively `personalization.' 
Our approach is inspired by the standard practice of mean image subtraction. To remove appearance based variation, we construct a mean neutral image, one for each person/session pair by averaging together first $5$ seconds of their neutral images,
as neutral images are stable over time. 
This image is subtracted from all other images derived from the originating person/session pair to effectively normalize it for a user (see \fig{fig:personalization}) per Equation~\ref{eq:personal} where $P(I)$ is the CNN test input, $I$ is the original image, and $N_u$ is the set of neutral images for that user. An `in practice' realization of this technique would be requesting a user maintain a minimally emotive expression for a short period of time and using the accumulated data to construct a similar normalizing image effectively giving each person a different mean subtraction. This generates $P(I)$ for new/test users.

\begin{equation}
	P(I) = I - \frac{1}{\left\vert{N_u}\right\vert}\sum_{I_u \in N_u}{I_u}
	\label{eq:personal}
\end{equation}

Repeated experiments demonstrated that `personalizing' images by subtracting a separate mean image per user is an effective means of increasing the accuracy of the classifier. 

To avoid introducing a bias towards statistical significance when testing the effectiveness of personalization, we first collapsed across results within a single subject--session--condition triplet (\eg, subject 28, session 2, condition `brow raise') into a single average accuracy and $F1$ score, then conducted a paired 1-tailed $t$-test to test the significance of the difference in average accuracy across the 23 users between results obtained with and without `personalization.' 
Action unit expression classification was significantly increased by the introduction of personalization ($p < 0.001$) as was `emotive' expression classification ($p = 0.018$).

%% file: results.tex

\begin{table}[b!]
\begin{center}
\begin{tabular}{ |l|c| }
\hline
\multicolumn{1}{|c|}{\textbf{Method}} & \textbf{Accuracy}\\
\hline
1: Average of human raters & 47\%\\
2: Average of human raters w/neutral & 50\%\\
3: Advanced human rater & 58\%\\
4: Advanced human raters w/neutral & 62\%\\
\hline
5: Variants based off of \cite{cvpr2016_Khosla} & 26\%\\
6: InceptionV3 one tower per eye & 48\%\\
7: InceptionV3 w/ frozen weights & 55\%\\
\hline
8: InceptionV3 HMD1 & 65\%\\
9: InceptionV3 HMD2 & 69\%\\
\hline
10: \textbf{Eyemotion HMD2} & \textbf{73}\%\\
\hline
\end{tabular}
\end{center}
\caption{\label{table:ablation} Preliminary tests using 4 left out participants.} 

\end{table}

\section{Experiments \& Results}\label{sec:results}

We experiment with benchmarks of different convolutional neural networks in a small, single validation set study on the 10 action units listed in Section~\ref{sec:data} with 4 holdout participants and 19 participants to train on from HMD1.
This yields a comparison metric for our proposed architecture. The different architectures we tried as benchmarks are listed in Table \ref{table:ablation}. For all our uses of InceptionV3, we fine-tune a model trained off of the ImageNet dataset~\cite{ILSVRC15}. 

Methods 1-4 were results from human raters as described in Section \ref{sec:human_accuracy}. Method 5 is the result obtained from a two tower (one-per-eye) gaze tracking network adapted from~\cite{cvpr2016_Khosla}. Method 6 is InceptionV3 with two towers (one for each eye) and tied weights. Method 7 is a single-tower InceptionV3, which receives concatenated eye images, but only the last fully connected softmax layers are allowed to vary. Methods 8 and 9 are fine-tuned InceptionV3 with data augmentation as described in Section \ref{sec:data_aug}. Method 10 is our Eyemotion approach detailed in Section~\ref{sec:approach}.

Initial tests we performed on different participants showed that results across individuals (without personalization) vary drastically and range between 
31\% to 90\% between users/headsets (see Supplementary materials).
Results depend on factors such as ergonomic fit, eyebrow color, eyebrow position, and expressiveness. Because of this, one hold out test set of participants is acceptable in a small study but not an effective enough performance measurement for comparison of our actual model accuracy. Therefore, in the experiments below, we use 5-fold cross validation by holding out participants on our dataset to get a more robust comparison between our approach and the baseline model. 

\begin{figure}[th!]
	\centering
	\includegraphics[width=0.4\textwidth]{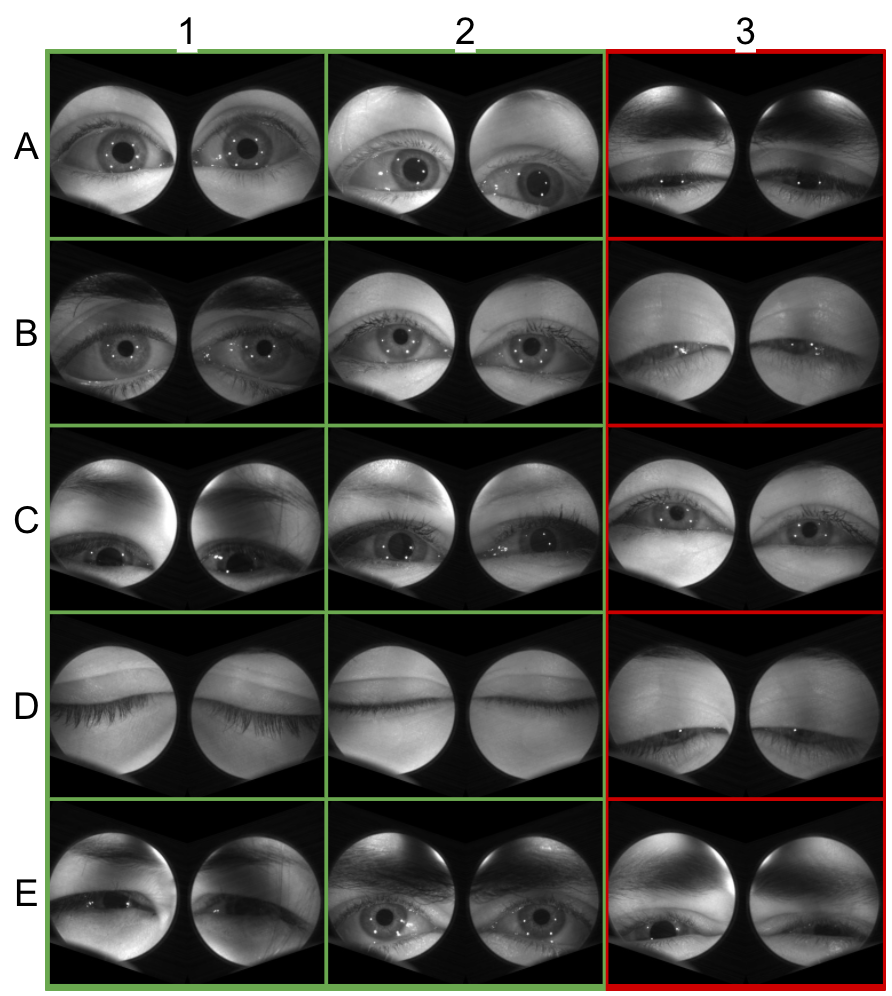}
	\caption{Visualizing the correct and incorrect classifications of the model for a subset of expressions. Columns 1 and 2 are correct classifications (green border), while column 3 are false negatives (red border). Rows are different expressions. A: Surprise, B: Neutral, C: Happiness, D: Eyes closed, E: Anger. In some cases, it is clear why the incorrect classification was made.}
	\label{fig:visualizing_results}
    \vspace{-.1in}
\end{figure}

We found results on HMD1 and HMD2 were roughly the same as shown in Table \ref{table:ablation}, so for brevity we present results for HMD2 only and include more results in our supplementary material. 
Our model achieves a mean accuracy on `emotive' expressions of 66.6\% and 73.7\% without and with personalization, respectively. Facial action unit classification accuracy was 63.7\% without personalization and 70.2\% with personalization (results in Fig~\ref{fig:visualizing_results}, per class details shown in Tables \ref{table:au_results}, \ref{table:au_personal_results}, \ref{table:expression_results}, and \ref{table:expression_personal_results}).  
Both results have higher than the average advanced human rater accuracy of 60.8\% discussed in Section \ref{sec:human_accuracy}. 

\begin{table}
\begin{center}
\begin{tabular}{ |c|c|c|c|c|  }
\hline
\textbf{Action Unit} & \textbf{Precision} & \textbf{Recall} & \textbf{$F_1$} & \textbf{Support}\\
\hline
\small{Brow lower} & 0.59 & 0.66 & 0.63 & 2897\\
\small{Upper lid raise} & 0.74 & 0.76 & 0.75 & 4912\\
\small{Cheek raise} & 0.56 & 0.64 & 0.60 & 2659\\
\small{Eyes closed} & 0.54 & 0.51 & 0.53 & 793\\
\small{Left brow raise} & 0.52 & 0.28 & 0.36 & 1598\\
\small{Left wink} & 0.95 & 0.86 & 0.90 & 2638\\
\small{Neutral} & 0.48 & 0.59 & 0.53 & 2309\\
\small{Right brow raise} & 0.29 & 0.18 & 0.22 & 1855\\
\small{Right wink} & 0.83 & 0.87 & 0.85 & 2220\\
\small{Squint} & 0.62 & 0.71 & 0.66 & 2135\\
\hline
Avg / total & 0.64 & 0.65 & 0.64 & 24016\\
\hline
\end{tabular}
\end{center}
\caption{Facial action units finetuned with InceptionV3. The overall mean accuracy 63.7\%.}
\label{table:au_results} 

\begin{center}
\begin{tabular}{ |c|c|c|c|c|  }
\hline
\textbf{Action Unit} & \textbf{Precision} & \textbf{Recall} & \textbf{$F_1$} & \textbf{Support}\\
\hline
\small{Brow lower} & 0.67 & 0.74 & 0.70 & 2576\\
\small{Upper lid raise} & 0.75 & 0.70 & 0.73 & 4956\\
\small{Cheek raise} & 0.65 & 0.64 & 0.64 & 2903\\
\small{Eyes closed} & 0.92 & 0.54 & 0.68 & 872\\
\small{Left brow raise} & 0.61 & 0.33 & 0.43 & 1616\\
\small{Left wink} & 0.93 & 0.90 & 0.91 & 2577\\
\small{Neutral} & 0.53 & 0.96 & 0.68 & 2372\\
\small{Right brow raise} & 0.50 & 0.24 & 0.32 & 1701\\
\small{Right wink} & 0.82 & 0.87 & 0.84 & 2235\\
\small{Squint} & 0.65 & 0.67 & 0.66 & 2208\\
\hline
Avg / total & 0.70 & 0.69 & 0.68 & 24016\\
\hline
\end{tabular}
\end{center}
\caption{Facial action units finetuned with InceptionV3 using our personalization method. Mean accuracy 70.2\%.}
\vspace{-.1in}
\label{table:au_personal_results} 
\end{table}
\begin{table}
\begin{center}
\begin{tabular}{ |c|c|c|c|c|  }
\hline
\textbf{Expression} & \textbf{Precision} & \textbf{Recall} & \textbf{$F_1$} & \textbf{Support}\\
\hline
Anger & 0.72 & 0.72 & 0.72 & 2695\\
Closed Eyes & 0.77 & 0.78 & 0.77 & 899\\
Happiness & 0.66 & 0.67 & 0.66 & 2610\\
Neutral & 0.60 & 0.78 & 0.68 & 2274\\
Surprise & 0.71 & 0.51 & 0.59 & 2425\\
\hline
Avg / total & 0.68 & 0.68 & 0.67 & 10903\\
\hline
\end{tabular}
\end{center}
\caption{Expressions finetuned with InceptionV3. Overall mean accuracy 66.6\%.}
\label{table:expression_results} 

\begin{center}
\begin{tabular}{ |c|c|c|c|c|  }
\hline
\textbf{Expression} & \textbf{Precision} & \textbf{Recall} & \textbf{$F_1$} & \textbf{Support}\\
\hline
Anger & 0.71 & 0.83 & 0.77 & 2455\\
Closed Eyes & 0.90 & 0.67 & 0.77 & 863\\
Happiness & 0.81 & 0.60 & 0.69 & 2644\\
Neutral & 0.64 & 0.95 & 0.77 & 2425\\
Surprise & 0.81 & 0.60 & 0.69 & 2516\\
\hline
Avg / total & 0.76 & 0.74 & 0.73 & 10903\\
\hline
\end{tabular}
\end{center}
\caption{Expressions finetuned with InceptionV3 using our personalization method. Overall mean accuracy 73.7\%.}
\label{table:expression_personal_results} 
\vspace{-.1in}
\end{table}

\subsection{Human Accuracy}\label{sec:human_accuracy}
We took a subset of 350 images of the data with different users and expressions in order to form a human benchmark of the data given that humans excel at pattern matching. We had 2 beginner classifiers (who had not seen any eye images previously), 1 intermediate classifier (who had seen some eye images previously), and 2 advanced, well-trained classifiers look at the concatenated eye images and guess the label from the 10 facial action units on each of those images.
We also allowed them to do this again but with a sample image of each user doing the neutral expression for comparison. This allows us to see if a personalization method can be useful for the classifier. The inter-rater kappa from our raters without and with a sample neutral image were $0.61$ and $0.64$ respectively showing good agreement among raters. The F1 score of the best rater was 0.63 without and 0.65 with neutral images (worse than our method's results). This suggests that some amount of one-shot learned personalization should be able to improve results.

\subsection{Applications}\label{sec:applications}

\begin{figure*}[t] 
\begin{center}
\begin{subfigure}{0.28\textwidth}
	\includegraphics[width=\textwidth]{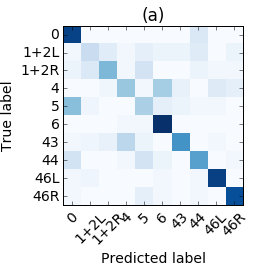}
	\label{fig:advanced_human}
\end{subfigure}
~
\begin{subfigure}{0.28\textwidth}	\includegraphics[width=\textwidth]{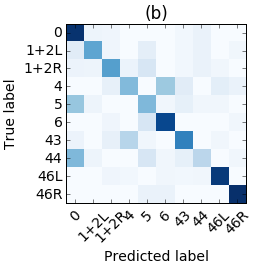}
	\label{fig:advanced_human_wneutral}
\end{subfigure}
~
\begin{subfigure}{0.33\textwidth}
	\includegraphics[width=\textwidth]{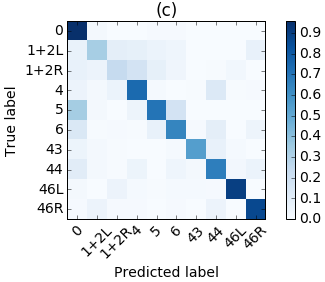}
    \label{fig:cv_users}
\end{subfigure}
\vspace{-.1in}
\caption{Confusion matrices of advanced human raters showing the accuracy of a)Advanced human, b) Advanced human with a given neutral image, c) Our network tested on holdout users never seen before. Labels are \textsl{Neutral (AU0), Left Brow Raise (AU1+2L), Right Brow Raise (AU1+2R), Brow Lower (AU4), Upper Lid Raise (AU5), Squint (AU44), Both Eyes Closed (AU43), Left Wink (AU46L), Right Wink (AU46R), and Cheek Raise (AU6)}.}
\end{center}
\vspace{-.1in}
\end{figure*}

Given our model produces a probability distribution of AUs or expressions, we can use that output to infer what a user is feeling and/or drive an avatar showing their expressions in real-time. Smoothing with exponential decay is applied on the inferred expression probabilities to get temporally stable labels. The model also yields a more intuitive expression medium for VR as opposed to gestures or keyboard inputs.
We can also adjust a user's VR environment based on their expression and hence their inferred emotional state.

\begin{figure}[ht]
\begin{center}
\begin{subfigure}{0.22\textwidth}
	\centering
	\includegraphics[width=0.9\textwidth]{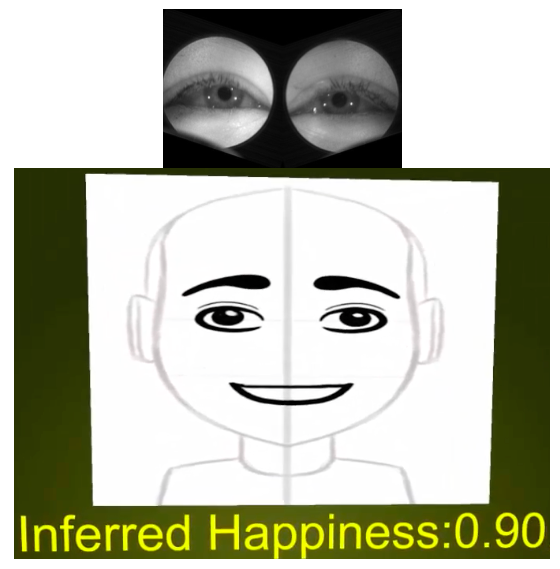}
	\caption{Animation still with our model inferring happiness.}
	\label{fig:infer_happy}
\end{subfigure}
\begin{subfigure}{0.22\textwidth}
\centering
\includegraphics[width=0.75\textwidth]{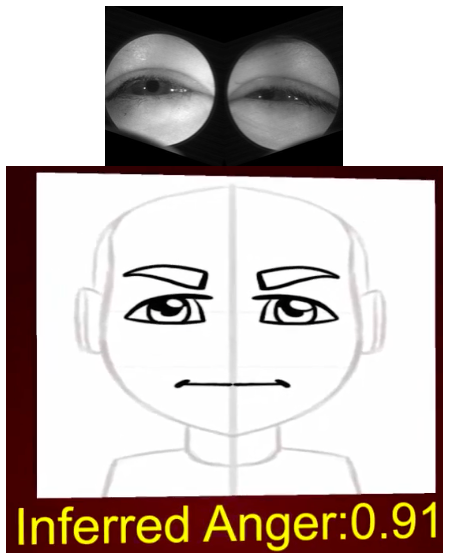}
	\caption{Animation still with our model inferring anger.}
	\label{fig:infer_anger}
\end{subfigure}
\caption{An animation that can be driven by our model's inference.}
\end{center}
\end{figure}

This can be demonstrated with a simple example where we change the ambient light and color based on the user's expressions while they watch a film clip as shown in the supplementary material. An example of this is included in the supplementary video, and shown in 
~\fig{fig:infer_happy} and \fig{fig:infer_anger}. Clearly these are very simple applications of our framework, and far more elaborate adjustments of the ambient environment are possible.

%% file: discussion.tex
\section{Discussion}

Our initial studies rapidly demonstrated the superiority of the concatenated-eye architecture, in which eye images are concatenated together before being input ({\em see} Table \ref{table:ablation}).
This architecture outperformed all others by ten percentage points or more when evaluated on the hold-out participants. We hypothesize that the improvement is due to concatenating the eye images at the outset, giving the CNN the ability to combine information from each eye. When the testing was extended to four other folds, such that each participant is excluded from training in exactly one fold, this accuracy improvement remained. 
Introducing a novel eye personalization technique, motivated by mean image subtraction, further improved the accuracy by nearly five percentage points  on average, which $t$-tests showed was significant (all $p < 0.05$). Typical errors occured when a visible component (\eg eyebrows) was occluded ({\em see}  \fig{fig:visualizing_results}).

We compare our personalization method to nonpersonalization methods and an implementation inspired by \cite{cvpr2016_Khosla}. Compared to other attempts at personalization, we instead explicitly create a difference from the mean neutral image and show this yields better results on average across all classes and especially improving neutral classification.
In addition, some comparison could be made to \cite{masai2016facial} where they classify 8 expressions on 8 users achieving good accuracy on holdout sessions. However, on hold-out users, they achieve an accuracy of at most 48\% and determined that a new user requires individual training for classification. Our deep learning personalization method scales well to new users and requires no individual training.

We also found that our implementation is robust to different hair colors and makeup. However, incorrect labels during data collection harm our results, which is why left brow raise and right brow raise have such low precision and recall as many people cannot do both and, critically, participants during data collection would attempt to do them even if they could not. This is a downside of autonomous data collection in which participants are asked to perform, as we have no means other than the participant's self report that the performances are accurate. It is, however, non-trivial to collect in-the-wild or unoccluded data for this task.

\paragraph{Limitations and Future Work:}
Data collection is an involved process and requires that participants produce and sustain expressions effectively. Even a model that can learn to recognize these synthetic expressions arbitrarily well may learn only to recognize what is essentially a biased facsimile of a real expression. This is because our participants may be instructed to, for instance, {\em act} angry but cannot be instructed to {\em be} angry. For this reason, collecting ground truth data is difficult. Further, this work makes a simplifying assumption that expressions are discrete: they are either occurring or not occurring and never in a transition state. It also assumes that only one expression can occur at a time. Since none of these assumptions are actually true, our model cannot accurately capture the breadth of facial expressiveness even if it had infinitely many samples.

Future work is inspired by our limitations. Using naturalistic stimuli (such as video) and treating the task as a clustering or an alignment would spare participants the requirement of having to `fake' expressions and remove the requirement of imposing a fixed set of expressions. We would also like to verify the accuracy of our collected labels with expert FACS raters.

%% file: closing.tex
\section{Conclusions}

A large amount of information about facial expressions is encoded in the eyes alone. Perhaps more surprising, this work shows that such information is interpretable by a CNN and can automate expression inference, even when the images are relatively low fidelity and derived from consumer-grade equipment. Even with 50 users, our model handily exceeds human classification accuracy and can power an avatar that serves as an expressive surrogate for a user wearing an HMD. We have demonstrated using consumer-grade eye-tracking cameras, which are already being included in VR headsets, a means to preserve and transmit social information among users engaged in VR. \emph{We also propose a novel personalization method with no retraining that achieves 74\% accuracy on classifying expressions and 70\% accuracy on classifying facial action units.}

\section*{Acknowledgments} 

We thank our collaborators in Google Machine Perception Research and Daydream Labs, and in particular, Hayes Raffle, Alex Wong, Sergio Guadarrama, Brendan Jou, Chris Bregler, and Sergey Ioffe for their insightful suggestions and support.